%% file: main.tex
\documentclass[10pt,twocolumn,letterpaper]{article}

\usepackage[pagenumbers]{cvpr} 

\usepackage{multirow}
\usepackage{booktabs}
\usepackage{appendix}
\usepackage{algorithm}
\usepackage{algpseudocode}

\title{A Scalable Attention-Based Approach for Image-to-3D Texture Mapping}

\author{Arianna Rampini\\
Autodesk Research\\
Milan, IT\\
{\tt\small arianna.rampini@autodesk.com}
\and
Kanika Madan\\
Autodesk Research\\
Toronto, CA\\
{\tt\small arianna.rampini@autodesk.com}
\and
Bruno Roy\\
Autodesk Research\\
Montreal, CA\\
{\tt\small arianna.rampini@autodesk.com}
\and
AmirHossein Zamani\\
Mila, Concordia University, and Autodesk Research\\
Montreal, CA\\
{\tt\small amirhossein.zamani@mila.quebec}
\and
Derek Cheung\\
Autodesk Research\\
Toronto, CA\\
{\tt\small derek.cheung@autodesk.com}
}

\begin{document}
\maketitle



\input{Sections/0_abstract}

\input{Sections/1_intro}

\input{Sections/2_related_work}
\input{Sections/3_method}
\input{Sections/4_results}

\input{Sections/5_conclusion}

\small
\bibliographystyle{ieeenat_fullname}
\bibliography{main_bib}

\end{document}

%% file: Sections/0_abstract.tex
\begin{abstract}

High-quality textures are critical for realistic 3D content creation, yet existing generative methods are slow, rely on UV maps, and often fail to remain faithful to a reference image. To address these challenges, we propose a transformer-based framework that predicts a 3D texture field directly from a single image and a mesh, eliminating the need for UV mapping and differentiable rendering, and enabling faster texture generation. Our method integrates a triplane representation with depth-based backprojection losses, enabling efficient training and faster inference. Once trained, it generates high-fidelity textures in a single forward pass, requiring only $\sim$0.2s per shape. Extensive qualitative, quantitative, and user preference evaluations demonstrate that our method outperforms state-of-the-art baselines on single-image texture reconstruction in terms of both fidelity to the input image and perceptual quality, highlighting its practicality for scalable, high-quality, and controllable 3D content creation.

\end{abstract}

%% file: Sections/1_intro.tex
\section{Introduction}
\label{sec:intro}

3D content creation is central to applications in gaming, virtual and augmented reality, digital twins, and immersive media. With the rapid progress of generative models for 3D shapes, it is now possible to automatically generate diverse and detailed geometries, greatly accelerating the creative process for designers and developers. To be practically useful in many applications, 3D shapes also require high-quality textures that are faithful to a given input reference, enabling realistic appearance and stylistic control. Generating such textures remains a fundamental challenge in computer vision and computer graphics.

\begin{figure}
    \includegraphics[width=\linewidth]{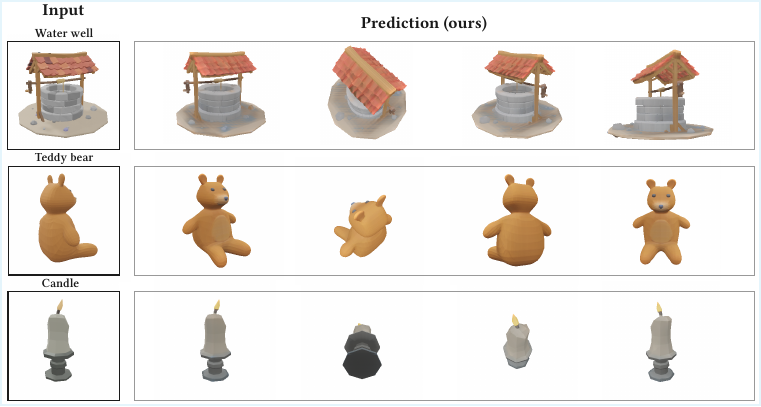}
    \centering
    \caption{Given a single input image (left), our method predicts a texture field for the corresponding 3D mesh and generates high-fidelity textures in a single forward pass. The figure shows novel views of textured meshes produced by our approach for various objects.}
    \label{fig:teaser}
\end{figure}

Existing approaches for texture generation have achieved impressive visual results, particularly when relying on image diffusion models \cite{DiffusionModels, StableDiffusionModel} or multi-view rendering pipelines \cite{chen2023text2tex, zeng2023paint3d}. These methods are capable of producing high-resolution textures, but they share several important limitations that restrict their use in practice. First, they are computationally expensive, often requiring several minutes per object due to iterative optimization or multi-view rendering. Second, they typically assume access to a clean mesh with a predefined UV mapping, which is rarely available for outputs from modern generative models or real-world 3D scans. Third, despite their high resolution, the generated textures are not always faithful to a reference input image, limiting their applicability in reconstruction or editing tasks where accuracy is crucial.

In this work, we introduce a transformer-based framework that directly predicts a 3D texture field from a single input image and a mesh, without the need for UV mapping, making it robust to noisy or incomplete geometry. Our method represents textures through a triplane field and computes supervision via depth-based backprojection, enabling efficient training. Although individual components—transformer backbones, triplane representations, and differentiable projections—are well established, their integration yields a model that achieves a strong balance between speed, flexibility, and quality. Once trained, our approach produces high-fidelity textures in a single forward pass, requiring only a fraction of a second ($\sim 0.2$s) per shape, compared to several minutes for existing methods.

We validate our method extensively on single-image texture reconstruction benchmarks, where it consistently outperforms state-of-the-art baselines both in fidelity to the input image and overall texture quality. Furthermore, we conduct a user study, which confirms that human evaluators strongly prefer the textures generated by our approach over those from competing methods. These results make our model particularly well-suited for modern 3D content creation pipelines, where scalability, speed, and accuracy are critical. Our contributions can be summarized as follows:

\begin{enumerate}
\item We propose a simple yet effective transformer-based framework for predicting texture fields from a single image and a mesh, without requiring UV mapping or multi-view rendering.
\item We integrate a triplane texture representation with depth-based backprojection losses, enabling efficient and scalable training.
\item Our method generates high-fidelity textures in a single forward pass ($\sim 0.2$s per shape), substantially faster than existing baselines that require minutes per object.
\item We show strong improvements over state-of-the-art methods on single-image texture reconstruction tasks, supported both by quantitative metrics and by a user study.
\end{enumerate}

%% file: Sections/2_related_work.tex
\section{Related Work}
\label{sec:related_work}

Existing approaches to texture generation are broadly classified into the following categories.

\subsection{Directly using 3D Data}
\label{subsec:related_work_3D}


One of the earliest methods to learn texture as a continuous 3D function was TextureFields \cite{oechsle2019texturefields}, which learns an implicit texture representation that can predict the color for any 3D point on a shape. While flexible, continuous texture fields can struggle to reproduce very high-frequency surface detail compared to explicit UV maps. Other approaches learn to generate texture directly on mesh surfaces using convolutional or neural-field–based operators. Texturify \cite{siddiqui2022texturify} trains a GAN-style model to generate geometry-aware surface textures from collections of untextured shapes and images, and Mesh2Tex \cite{bokhovkin2023mesh2tex} learns a hybrid mesh–neural-field texture manifold that maps image queries to compact, high-resolution textures for a given mesh. These mesh-based generators often leverage adversarial training (GANs \cite{goodfellow2014gans} or StyleGAN \cite{karras2019stylegan} variants) to improve realism but inherit GAN failure issues such as mode collapse and training instability. 

More recently, diffusion- and point-cloud–based approaches have emerged to better capture local detail and to operate directly in 3D or UV space. TUVF \cite{cheng2023tuvf} learns generalizable UV radiance fields that disentangle texture from geometry by generating textures in a canonical UV sphere space. Point-UV \cite{yu2023pointuv} and related point-based diffusion pipelines produce coarse-to-fine textures by denoising colored point samples and then projecting them to UV maps. TexOct \cite{liu2024texoct} proposes an octree-based 3D diffusion to generate textures directly in 3D space, alleviating occlusion and sparse-sampling issues present in some point-based pipelines. While these methods improve high-frequency detail and 3D consistency, they are often demonstrated on limited datasets or category-specific collections, which makes broad generalization and scaling to many categories challenging.

\subsection{Multiview-Based Generation}
\label{subsec:related_work_multiview}


A different line of work targets using multi-view images to generate 3D textures that are consistent across views. Early iterative view-by-view inpainting approaches such as TEXTure \cite{richardson2023texture}, Text2Tex \cite{chen2023text2tex}, and InTeX \cite{InTeX}  generate colors for a mesh by repeatedly rendering the object from different viewpoints and inpainting or updating the visible texels in each view. However, such iterative, view-sequential procedures may produce inconsistencies across views because they lack global 3D awareness of the surface and lacks a deep understanding of 3D structure, often misaligning textures with task needs. 

To alleviate these issues, follow-up methods introduced several strategies. \cite{TextureRewardLearning} builds on this paradigm by introducing a geometry-aware fine-tuning stage that aligns textures with human preferences and task-specific objectives, improving coherence and control without the overhead of joint optimization of geometry and appearance. TexFusion \cite{cao2023texfusion} interleaves texture synthesis with multi-view denoising steps and performs view-consistent diffusion sampling to reduce per-view artifacts and stitching errors. TexPainter  \cite{zhang2024texpainter} enforces multi-view consistency by fusing latent views into a common color-space texture and uses a color-fusion optimization scheme (together with synchronized multi-view denoising) to reduce inconsistencies. Paint3D \cite{zeng2023paint3d} addresses lighting and baked-shading artifacts by separating coarse multi-view fusion from a learned UV inpainting / UVHD refinement stage, producing lighting-less high-resolution UV maps suitable for relighting, though it still relies on expensive test-time refinement. TexGen \cite{TexGen} further improves view consistency and detail preservation by introducing an attention-guided multi-view sampling and noise-resampling framework that maintains a time-dependent texture map updated across denoising steps, reducing seams and preserving fine appearance details.

\begin{figure*}[t]
    \centering
    \includegraphics[width=0.72\textwidth]{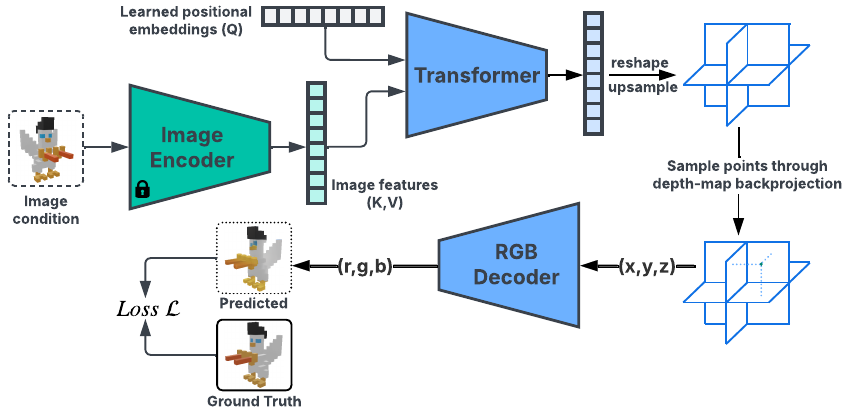}
    \caption{An overview of the training stage of our method. Given a single input image and a 3D mesh, we extract visual features from the image using a pre-trained DINO \cite{dinov2} encoder. Learned positional embeddings are processed by a transformer and fused with the visual features through cross-attention. The output is reshaped into a triplane texture representation. Query points sampled via depth-map backprojection are decoded into RGB values, yielding a 3D texture field. The model is trained end-to-end with supervision from ground-truth colors.}
    \label{fig:method_figure}
\end{figure*}

\subsection{Optimization-Based Methods}
\label{subsec:related_work_optimization}


These methods treat 3D shape parameters as learnable and leverages CLIP \cite{radford2021clip} and text-to-image or image-to-image diffusion models \cite{Imagen, DiffusionModels, StableDiffusionModel} in the form of Score Distillation Sampling (SDS) \cite{poole2022dreamfusion} as supervision. Early methods optimize mesh color or neural style fields directly using CLIP \cite{radford2021clip} losses to align renderings with text prompts, examples include \cite{chen2022tango,hong2022avatarclip,ma2023xmesh,michel2021text2mesh,clipmesh} which stylize geometry and/or texture by differentiably rendering the asset and minimizing CLIP-based objectives. The introduction of SDS in \cite{poole2022dreamfusion} enabled a powerful new paradigm: pre-trained 2D text-to-image diffusion models can be used as priors/supervisors to optimize 3D representations (e.g., NeRFs \cite{mildenhall2020nerf} or sparse hash grids \cite{instantngp}) by distilling their score into a differentiable 3D objective. DreamFusion \cite{poole2022dreamfusion} is the canonical example of this approach and inspired many follow-ups including Magic3D \cite{lin2023magic3d} and ProlificDreamer \cite{wang2023prolificdreamer}). Subsequent work extended SDS-based optimization to produce better geometry–appearance disentanglement and higher-fidelity materials. Fantasia3D \cite{chen2023fantasia3d} and other hybrid pipelines disentangle geometry and appearance and introduce spatially-varying BRDF/material representations during optimization, while TextureDreamer \cite{yeh2024texturedreamer} and DreamMat \cite{zhang2024dreammat} incorporate geometry- and light-aware diffusion objectives to improve relightable texture and PBR material estimation. 

Despite these advances, optimization-based methods still suffer from practical shortcomings: (i) they can be computationally expensive (per-scene optimization that takes minutes to hours), (ii) they may produce view-inconsistent artifacts or “Janus” faces without careful debiasing, and (iii) naive distillation from 2D models often leads to baked-in shading or incorrect material decomposition unless geometry- or light-aware priors are used. These limitations motivate hybrid, feed-forward, and geometry-aware texture objectives that explicitly inject geometric knowledge into texture synthesis.

\subsection{Feed-Forward Methods}
\label{subsec:related_work_ffmethods}


Recently, there has been a strong movement toward feedforward 3D generation models that are trained on large-scale data and produce high-quality 3D assets in a single forward pass, avoiding expensive per-object optimization. These methods typically adopt high-capacity transformer backbones and more compact 3D intermediate representations (e.g., triplanes, 3D Gaussian, or hybrid triplane-gaussian forms) to enable fast inference while maintaining competitive rendering quality. The large reconstruction model \cite{hong2024lrm} demonstrated that scaling model capacity and training on massive multi-view datasets enables generalizable single image to 3D reconstruction. LRM directly predicts a neural radiance field from an input image using a large transformer, producing robust reconstructions across many object categories. Instant3D \cite{li2023instant3d} showed that a carefully designed feedforward network can produce high-quality text-to-3D results in under one second by directly constructing a triplane representation from a text prompt, using mechanisms such as cross-attention and style injection to generate conditional language. Other works push the trade-offs between speed, generalization, and quality by changing the intermediate 3D primitive. GRM (Gaussian Reconstruction Model) \cite{xu2024grm} and related LRM variants represent scenes or objects as collections of 3D Gaussians decoded from image-aligned tokens. This enables extremely fast reconstructions while remaining amenable to transformer scaling and multi-view conditioning. 

Hybrid representations such as \cite{zou2023triplanemeetsgaussiansplatting} combine the best of both worlds. \cite{zou2023triplanemeetsgaussiansplatting} uses a point/triplane decoder to predict a hybrid triplane-gaussian intermediate, which is then rendered by fast splatting, resulting in better novel view rendering quality than naive explicit primitives while retaining the speed advantages of splatting-based renderers. These hybrid pipelines have proven effective for single-view reconstruction and fast feed-forward text/image to 3D tasks. 

Despite these advances, feed-forward approaches still face limitations relevant to texture generation: they often require large, diverse training datasets to learn high-frequency, material-aware texture priors; handling complex illumination and spatially varying BRDFs remains challenging; and many models prioritize geometric fidelity and rendering speed over fine, geometry-aware texture detail, which motivates methods that explicitly incorporate geometric cues (e.g., curvature-aware losses or texture alignment objectives) into the learning process.

%% file: Sections/3_method.tex
\section{Method}
\label{sec:method}


We present a novel approach for reconstructing high-quality textures on 3D shapes using a transformer-based architecture that synthesizes triplane representations. Our method learns a mapping from visual conditioning inputs to continuous texture fields represented as triplane features. An overview of our training and inference pipelines are shown in Fig.~\ref{fig:method_figure} and ~\cref {alg:inference}, respectively.

\subsection{Preliminaries}

A triplane representation is a 3D neural field encoding that decomposes a volumetric feature field into three orthogonal 2D feature planes corresponding to the $XY$, $XZ$, and $YZ$ coordinate planes. This representation, originally introduced in EG3D \cite{chan2021efficient}, provides an efficient way to represent continuous 3D features with 2D convolutional networks.

The Large Reconstruction Model (LRM) framework \cite{hong2024lrm} first introduced the combination of transformers with triplane representations for joint geometry and texture reconstruction using a NeRF field \cite{mildenhall2020nerf}. Building upon this architecture to tackle the more constrained problem of texture field reconstruction over known 3D meshes. Unlike LRM, which learns texture and geometry jointly through differentiable rendering and camera-view modulation, our approach focuses specifically on predicting texture fields over existing geometries. This introduces a unique correspondence problem: establishing the relationship between the (unknown) viewpoint of the conditioning image and the given 3D mesh.

Texture field prediction was pioneered in the TextureFields work \cite{oechsle2019texturefields}. However, that approach had limited scalability, being effective primarily on single-category ShapeNet data. Our contribution lies in combining the texture field rationale with the scalable triplane-transformer architecture, enabling texture synthesis across diverse object categories by training on large-scale datasets. Triplane representations provide both computational efficiency and representational power, allowing us to scale beyond single categories while maintaining the continuous texture field formulation.

Through our experiments, we found that explicit geometric encoding offered minimal benefit in our setting, leading us to adopt a streamlined architecture that relies only on visual conditioning. 

\subsection{Problem Formulation}

Given a 3D shape with known geometry and a conditioning image $I$, our goal is to learn a texture field
\[
T_\theta : \mathbb{R}^3 \to \mathbb{R}^3,
\]
that maps 3D coordinates to RGB colors. The texture field should be consistent with the visual appearance suggested by the conditioning image while respecting the underlying 3D geometry. We formalize this as:
\[
\text{RGB} = T_\theta(p, I),
\]
where $p \in \mathbb{R}^3$ represents 3D coordinates and $I$ is the input image.

\subsection{Architecture}

\paragraph{Visual Conditioning.} We employ a DINOv2 encoder \cite{dinov2} that processes RGB conditioning images at $384 \times 384$ resolution, producing visual features $z \in \mathbb{R}^{768}$ that capture semantic and appearance information.
 
\paragraph{Transformer-based Triplane Decoder.}  
In our implementation, the triplane consists of three feature maps of dimensions $[f_{dim}, t_{res}, t_{res}]$, where $f_{dim}=48$ is the feature dimension, $t_{res} \times t_{res}$ is the spatial resolution, the three planes correspond to orthogonal projections $XY$, $XZ$, and $YZ$.

The transformer decoder processes learned positional embeddings $f_{\text{init}}$ corresponding to triplane token positions. The learned positional embeddings are initialized using sinusoidal encoding and correspond to the flattened sequence of triplane tokens ($3072 = 3 \times t_{res}^2$ positions). The embeddings have the same dimensionality of the transformer hidden size, and are optimized end-to-end with the rest of the model.  Visual conditioning is integrated through cross-attention mechanisms in each transformer layer:
\[
f_{\text{out}} = \texttt{TransformerDecoder}(f_{\text{init}}, z).
\]

The transformer outputs are reshaped into spatial triplane format. Starting from $32 \times 32$ resolution, a convolutional upsampling network generates triplane features 
$P \in \mathbb{R}^{3 \times 48 \times 64 \times 64}$.

To sample features at an arbitrary 3D point $p = (x,y,z)$, we:
\begin{enumerate}
    \item Project $p$ onto each of the three planes;
    \item Use bilinear interpolation to sample features from each plane at the projected coordinates, yielding $f_{xy}$, $f_{xz}$, and $f_{yz}$;
    \item Concatenate the sampled features, producing a 144-dimensional feature vector ($3 \times 48$).
\end{enumerate}

Finally, the concatenated feature vector is passed through a 4-layer MLP (\texttt{RGBDecoder}) with ReLU activations to predict the final RGB color:
\[
\text{RGB}(p) = \texttt{RGBDecoder}([f_{xy}, f_{xz}, f_{yz}]).
\]

At inference time, our approach requires only a single forward pass of the pipeline, as summarized in Algorithm~\ref{alg:inference}.



\subsection{Training Methodology}

We train the model using precomputed multi-view depth maps with corresponding ground-truth images. Depth maps are converted into 3D point coordinates via backprojection with camera intrinsics as detailed below, and the predicted colors are supervised against ground truth. For each training sample, we process 4 random views from a set of 55 precomputed depth maps to ensure broad appearance coverage.

\paragraph{Depth-Map Backprojection.}  
To supervise the predicted texture field, we backproject depth maps into 3D point clouds, associating each 3D point with its ground-truth RGB value from the corresponding view.  

Formally, given a pixel $(u,v)$ with depth $d(u,v)$, its camera-space coordinate is obtained via inverse projection:
\[
\mathbf{p}_{c} = d(u,v) \, K^{-1} 
\begin{bmatrix} u \\ v \\ 1 \end{bmatrix},
\]
where $K$ is the camera intrinsics matrix.  
The point is then transformed into world coordinates using the camera-to-world transformation $T_{\text{cam}}$:  
\[
\mathbf{p}_{w} = T_{\text{cam}} \, \begin{bmatrix} \mathbf{p}_{c} \\ 1 \end{bmatrix}.
\]

Each depth map of resolution $384 \times 384$ yields $147{,}456$ points, one per pixel. We sample $4$ views per training step, leading to approximately $590$K points per batch, as background pixels are masked out and excluded from supervision.  
The reconstructed 3D points are queried into the predicted texture field to obtain a predicted image, which is then compared against the ground-truth one in the loss.

\paragraph{Loss Function.}  
We optimize a combination of a pixel-wise reconstruction loss and a perceptual similarity loss:  
\[
\mathcal{L}_{\text{total}} = \lambda_{\text{pixel}} \mathcal{L}_{\text{pixel}} + \lambda_{\text{perc}} \mathcal{L}_{\text{perc}}.
\]

The pixel-level term enforces low-level fidelity by directly comparing RGB values between predicted and ground-truth renderings:  
\[
\mathcal{L}_{\text{pixel}} = \frac{1}{V} \sum_{v=1}^V \| I_{\text{pred}}(v) - I_{\text{gt}}(v) \|_2^2,
\]
where $V$ denotes the number of views per sample.  

The perceptual loss $\mathcal{L}_{\text{LPIPS}}$ encourages high-level similarity by comparing features extracted from a pre-trained VGG network~\cite{simonyan2014very}, following the LPIPS formulation~\cite{zhang2018unreasonable}. Instead of focusing on raw pixel differences, this loss aligns image representations in a deep feature space, improving texture realism and visual coherence:  
\[
\mathcal{L}_{\text{perc}} = \text{LPIPS}(I_{\text{pred}}, I_{\text{gt}}).
\]

In practice, we set both weights equally ($\lambda_{\text{pixel}} = \lambda_{\text{perc}} = 1$), which provided a good balance between preserving fine details and maintaining perceptual quality in our experiments.

\begin{algorithm}[t]
    \caption{Inference for texture reconstruction}
    \label{alg:inference}
    \begin{algorithmic}
        \State \textbf{Input:} Single conditioning image $I$, 3D mesh $M$
        \State Encode $I$ with \texttt{DINO} $\rightarrow z$
        \State Generate triplane features with \texttt{TransformerDecoder} $\rightarrow P$
        \For{each query point $p \in M$}
            \State Project $p$ onto $(x,y),(x,z),(y,z)$ planes
            \State Sample features via bilinear interpolation from $P$
            \State Concatenate features $\rightarrow f_{xyz}$
            \State Decode $f_{xyz}$ with \texttt{RGBDecoder} $\rightarrow c(p)$
        \EndFor
        \State \textbf{Output:} Textured mesh with predicted RGB field
    \end{algorithmic}
\end{algorithm}

%% file: Sections/4_results.tex
\section{Experimental Evaluation}
\label{sec:results}

\subsection{Setup}

\paragraph{Datasets.}
We train on \emph{Objaverse}~\cite{deitke2023objaverse}, a large-scale dataset of $\sim$800k textured 3D assets spanning diverse object categories.
We follow a 98\%/2\% train/validation split.
To evaluate cross-dataset generalization, we additionally test on the \emph{Google Scanned Objects (GSO)} benchmark~\cite{downs2022google}, consisting in real-world scanned meshes.
For both datasets, we precompute RGB images and corresponding depth maps at $384{\times}384$ resolution from 55 viewpoints.

\paragraph{Implementation details.}
We train all models using the AdamW optimizer with batch size $128$, weight decay $0.05$, and a cosine learning rate schedule (base LR $2{\times}10^{-4}$ with $10$k warmup steps).
Training is performed in mixed precision on a cluster of 32 NVIDIA A100 GPUs.
At inference, our method runs in $\sim$0.2\,s per mesh on a single NVIDIA A10 GPU.

\paragraph{Baselines.}
We compare against three recent image-guided texturing methods:
Paint3D~\cite{zeng2023paint3d}, EASI-Tex~\cite{perla2024easi}, and TEXTure~\cite{richardson2023texture}.
All methods rely on iterative optimization guided by multi-view diffusion models and assume a UV-mapped mesh as input.
We use the authors' official implementations and recommended hyperparameters, running each baseline to convergence. All methods are provided with the same conditioning image and mesh and evaluated on the same set of novel views. Note that for TEXTure~\cite{richardson2023texture}, the image-to-texture generation stage requires fine-tuning a diffusion model on the conditioning image, which is considerably more time-consuming than other baselines. To ensure feasibility when processing $100$ objects from the GSO dataset, we keep all hyperparameters as proposed by the authors, except that we reduce the parameter \textit{max train steps} from $10000$ to $1000$ to achieve practical computation time (i.e. $~20$ minutes).

\paragraph{Metrics.}
We evaluate texture reconstruction quality using three metrics:
\emph{CLIP-Score}~\cite{radford2021learning}, which measures semantic consistency between the conditioning image and rendered views in CLIP embedding space;
\emph{LPIPS}~\cite{lpips}, which assesses perceptual similarity between predicted and ground-truth novel views;
and \emph{PSNR}, a standard pixel-level reconstruction metric.
Higher CLIP and PSNR and lower LPIPS indicate better performance.

\begin{figure}[ht]
    \centering
    \includegraphics[width=1\linewidth]{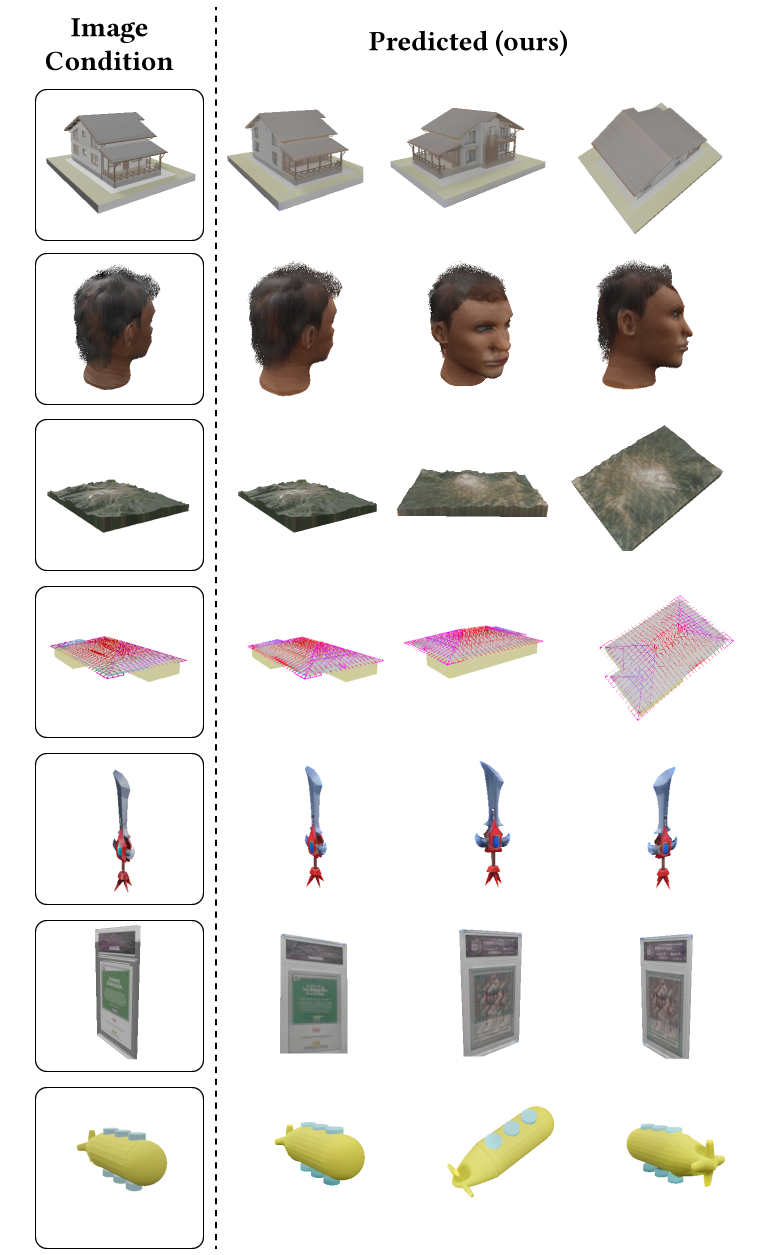}
    \caption{\textbf{Results on Objaverse (validation set).}
    Our feed-forward model generalizes across diverse categories and geometries, reconstructing high-fidelity textures from a single image (left).}
    \label{fig:additional_results}
\end{figure}


\subsection{Results}

We showcase examples of our texture reconstruction capabilities in Figures~\ref{fig:teaser} and \ref{fig:additional_results}.

\begin{figure}[hb]
    \centering
    \includegraphics[width=\linewidth]{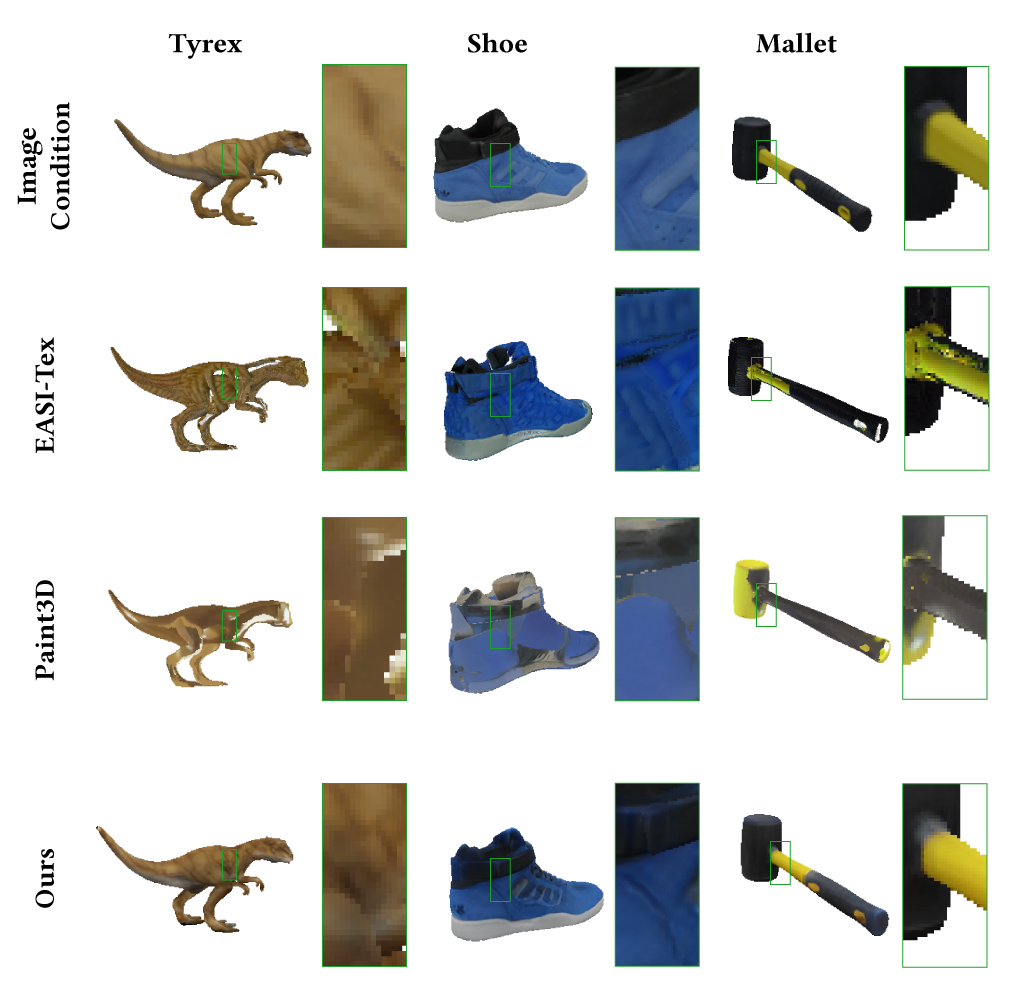}
    \caption{\textbf{Comparison on GSO.}
    Given the same conditioning image and mesh, our method (bottom row) produces textures with higher fidelity and fewer artifacts than diffusion-based baselines.}
    \label{fig:comparison}
\end{figure}

\begin{figure}[hb]
    \centering
    \includegraphics[width=\linewidth]{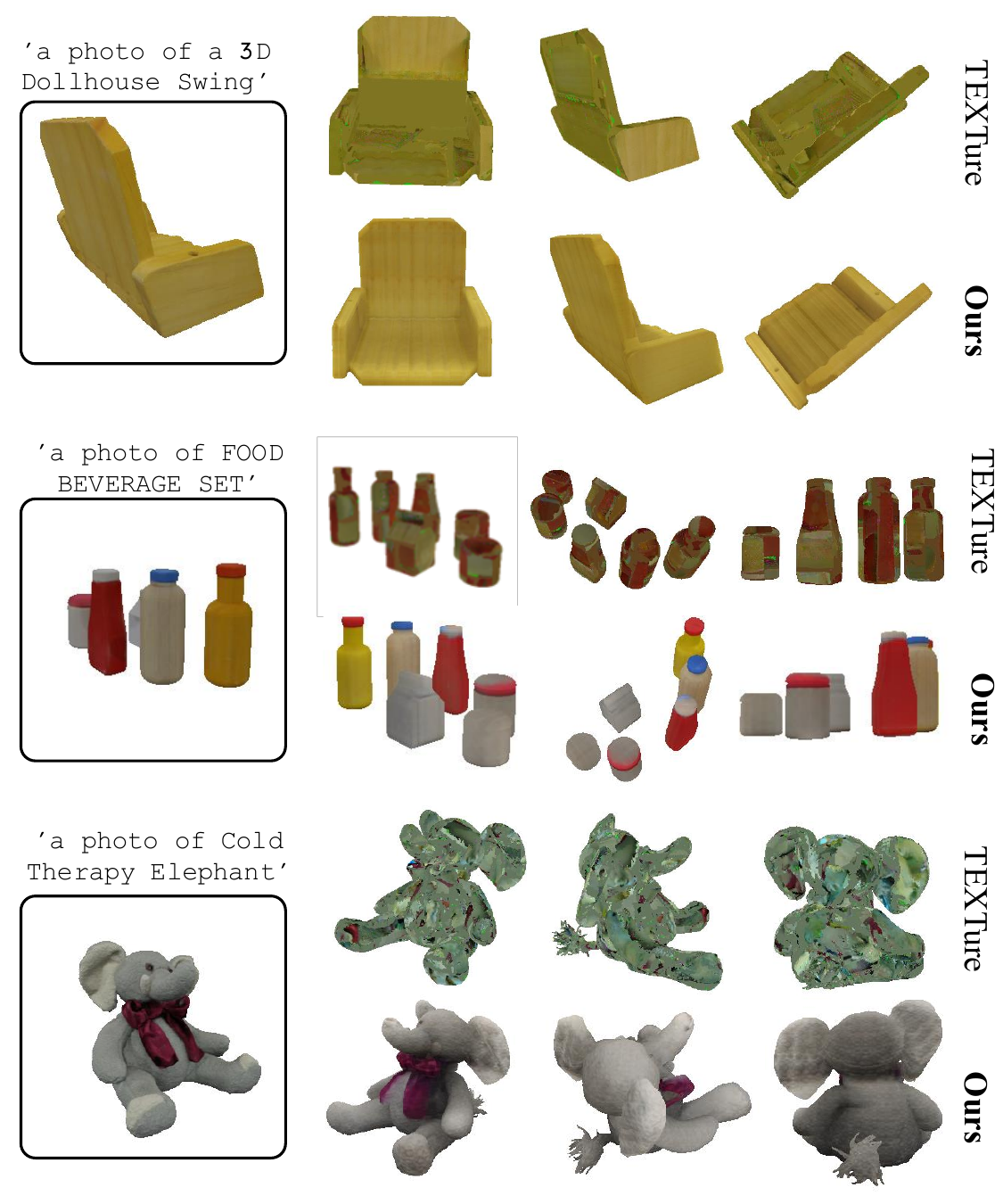}
    \caption{Comparison with TEXTure~\cite{richardson2023texture} on single-image texture reconstruction on GSO samples. TEXTure requires both a textual description and a conditioning image as input. Across varying levels of complexity—from simple uniform textures to multi-object scenes—our method generates coherent and faithful textures, whereas TEXTure often produces broken or inconsistent color patterns.}
    \label{fig:comparison2}
\end{figure}

\paragraph{Qualitative comparison.}
Figure~\ref{fig:comparison} presents close-up comparisons, while Figure~\ref{fig:comparison2} shows novel view generation.
We can observe that our method produces cleaner textures with fewer artifacts and significantly improved fidelity to the conditioning image compared to baselines.

\paragraph{Quantitative comparison.}
Table~\ref{tab:comparison} reports results on GSO (100 random objects, 10 novel views each).
Our approach outperforms both baselines across all metrics by a large margin.
Moreover, inference runs in only $\sim$0.2\,s per shape, which is orders of magnitude faster than optimization-based baselines (10--20 minutes).
Unlike prior work, our method requires no UV maps or mesh preprocessing, making it particularly suitable for pipelines that must handle hundreds of assets, such as procedural generation or large-scale 3D content creation.

\begin{table}[t]
\centering
\caption{\textbf{Comparison on GSO.} We evaluate 100 random objects and 10 novel views per object. Metrics are computed inside the silhouette. Our method achieves the best semantic alignment (CLIP), perceptual similarity (LPIPS), and pixel accuracy (PSNR) while running in $\sim$0.2\,s/shape without UV maps.}
\label{tab:comparison}
\begin{tabular}{l|ccc}
\toprule
\toprule
Method & CLIP-Score $\uparrow$ & LPIPS $\downarrow$ & PSNR $\uparrow$ \\
\midrule
TEXTure \cite{richardson2023texture} & 80.24 & 0.236 & 13.31 \\
Paint3D \cite{zeng2023paint3d} & 82.67 & 0.205 & 13.61 \\
EASI-Tex \cite{perla2024easi} & 83.25 & 0.203 & 13.72 \\
\textbf{Ours} & \textbf{90.09} & \textbf{0.075} & \textbf{27.65}\\
\bottomrule
\bottomrule
\end{tabular}
\end{table}

\subsection{Qualitative User Study}
We conducted a user study to evaluate and compare our method against two established baselines: Paint3D and EASI-Tex. The study involved 62 participants, all professionals working in the Media and Entertainment industry with a background in computer science. Participants were asked to assess the quality of the generated textures based on two key criteria: (1) realism and fidelity, and (2) consistency with respect to the condition image used to guide the generation. As summarized in Table~\ref{tab:evaluation_criteria}, our approach consistently outperformed both baselines across the evaluation metrics.

\begin{table}[h]
\small
\centering
\caption{A user study reports the percentage of participants who preferred the results from our approach across two evaluation criteria. Our method outperforms both Paint3D~\cite{zeng2023paint3d} and EASI-Tex~\cite{perla2024easi} in terms of texture fidelity and consistency relative to a conditional image.}
\label{tab:evaluation_criteria}
\begin{tabular}{l|ccc}
\toprule
\toprule
Evaluation Criteria & Paint3D & EASI-Tex & Ours\\
\toprule
Texture Realism \& Fidelity & 4.86 & 12.85 & \textbf{82.29} \\
Conditional Consistency & 0.34 & 3.46 & \textbf{96.21} \\
\bottomrule
\bottomrule
\end{tabular}
\end{table}

\begin{table*}[h!]
\centering
\caption{\textbf{Ablation on model size.} Validation is computed on Objaverse (val set); GSO is out-of-domain test. Best and second-best are in \textbf{bold} and \textit{italic}. The \textsc{base} model attains near-optimal performance at a significantly lower cost than \textsc{large}.}
\label{tab:ablation}
\begin{tabular}{l|cccc|cccc}
\toprule
\toprule
\multirow{2}{*}{Models}  & \multicolumn{4}{c}{Validation dataset} & \multicolumn{4}{c}{GSO (Test dataset)} \\
\cline{2-9}
& CLIP-Score $\uparrow$ & LPIPS $\downarrow$ & PSNR $\uparrow$ & MSE $\downarrow$ & CLIP-Score $\uparrow$ & LPIPS $\downarrow$ & PSNR $\uparrow$ & MSE $\downarrow$ \\
\midrule
\textsc{small}      & 89.2 & 0.066 & 23.52 & 0.110 & 84.8 & 0.091 & 23.29 & 0.088 \\
\textsc{base}       & \textbf{90.8} & \textit{0.051} & \textit{25.38} & \textbf{0.073} & \textit{88.3} & \textbf{0.071} & \textbf{25.93} & \textbf{0.047} \\
\textsc{large}        & \textbf{90.8} & \textbf{0.050} & \textbf{25.47} & \textbf{0.073} & \textbf{88.6} & \textbf{0.071} & \textit{25.78} & \textit{0.050} \\
\bottomrule
\bottomrule
\end{tabular}
\vspace{-5pt}
\end{table*}

\subsection{Ablations}
\label{sec:ablations}

\paragraph{Model capacity.}
We study the impact of model size by training three variants of our architecture:  
\textsc{small} ($\sim$9M parameters, 3 transformer layers, 6 attention heads, 384-dim features),  
\textsc{base} ($\sim$52M parameters, 9 layers, 9 heads, 576-dim features), and  
\textsc{large} ($\sim$115M parameters, 12 layers, 12 heads, 768-dim features).  
Results are reported in Table~\ref{tab:ablation}.  

Performance improves substantially from \textsc{small} to \textsc{base}, especially on perceptual metrics (LPIPS and CLIP).  
Increasing to \textsc{large} provides only marginal improvements ($<0.2$ CLIP, $<0.1$ PSNR) at the cost of more than doubling the number of parameters and training/inference memory usage.  
This suggests that texture reconstruction benefits from a moderately deep transformer with sufficient feature dimensionality, but quickly saturates as capacity grows.  
In practice, the \textsc{base} model offers the best trade-off between quality and efficiency, and is therefore used in all other experiments.  

\begin{figure}
    \centering
    \includegraphics[width=\linewidth]{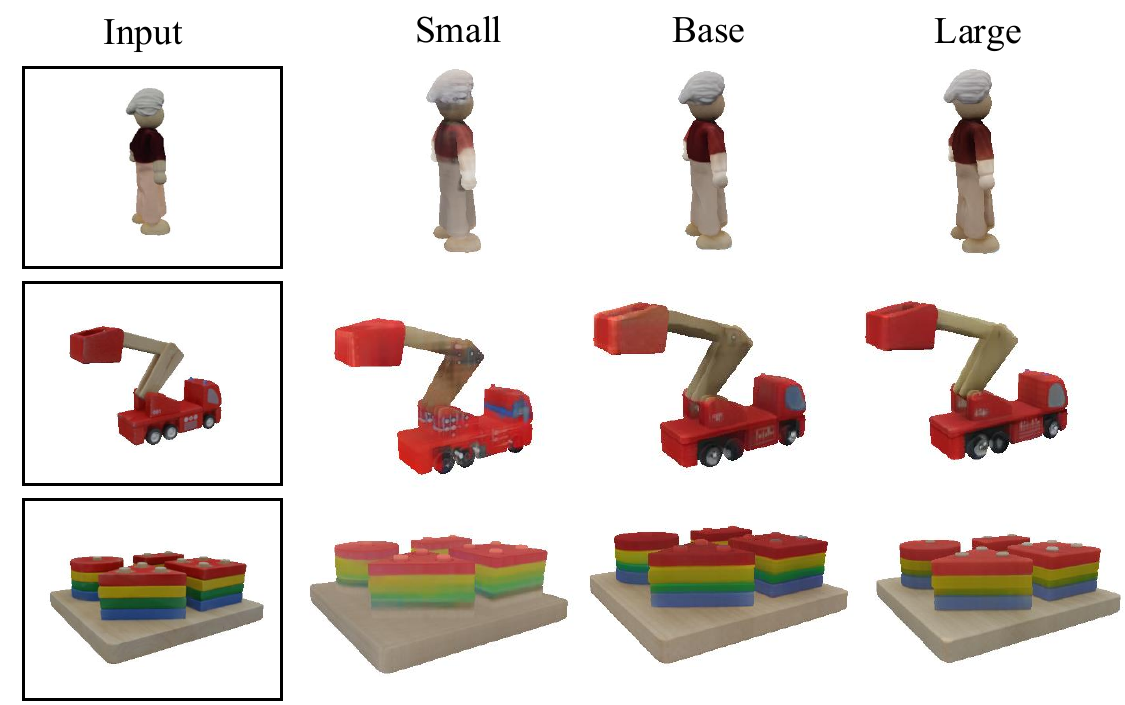}
    \caption{\textbf{Effect of model capacity.} Comparison of results from our \textsc{small}, \textsc{base}, and \textsc{large} variants. Larger models improve texture sharpness and color fidelity, especially for fine-grained structures, though the \textsc{base} model already provides a strong balance between quality and efficiency. \vspace{-10pt}}
    \label{fig:placeholder}
\end{figure}

\paragraph{Geometric conditioning.}
We experimented with adding geometric signals via cross-attention: (i) latent features from a pre-trained SDF VQ-VAE (\textit{Latent})~\cite{sanghi2024wavelet}, and (ii) point-cloud features from PointNet~\cite{qi2017pointnet}.
As reported in Table~\ref{tab:cond_loss}, neither variant improved performance; both slightly degraded LPIPS/PSNR, likely due to misalignment noise and reduced capacity for appearance modeling.
We therefore omit geometric conditioning in our final model.

\paragraph{Perceptual loss.}
Removing LPIPS lowers MSE/PSNR trade-offs (slightly better MSE) but harms perceptual quality and semantic alignment (worse LPIPS and CLIP).
Consistent with qualitative observations, LPIPS guidance helps preserve fine appearance and avoids over-smoothing, so we retain it.





\begin{table}[h!]
\footnotesize
\centering
\caption{\textbf{Ablation on conditioning and losses (Objaverse val).} Geometric conditioning does not help in the single-image setting; while removing LPIPS harms perceptual quality.}
\label{tab:cond_loss}
\begin{tabular}{l|cccc}
\toprule
\toprule
\multirow{2}{*}{Model variant}  & \multicolumn{4}{c}{Validation dataset} \\
\cline{2-5}
& CLIP-Score $\uparrow$ & LPIPS $\downarrow$ & PSNR $\uparrow$ & MSE $\downarrow$ \\ 
\midrule
\textsc{base}        & \textbf{90.8} & \textbf{0.050} & \textbf{25.47} & \textit{0.073} \\ 
w/o LPIPS loss      & 88.6 & 0.071 & 25.89 & \textbf{0.062} \\ 
+ Latent cond.   & 90.5 & 0.053 & 25.01 & 0.076 \\ 
+ Point cloud cond.  & 90.2 & 0.057 & 24.84 & 0.081 \\ 
\bottomrule
\bottomrule
\end{tabular}
\vspace{-10pt}
\end{table}

\paragraph{Failure cases.}

\begin{figure}
    \centering
    \includegraphics[width=0.95\linewidth]{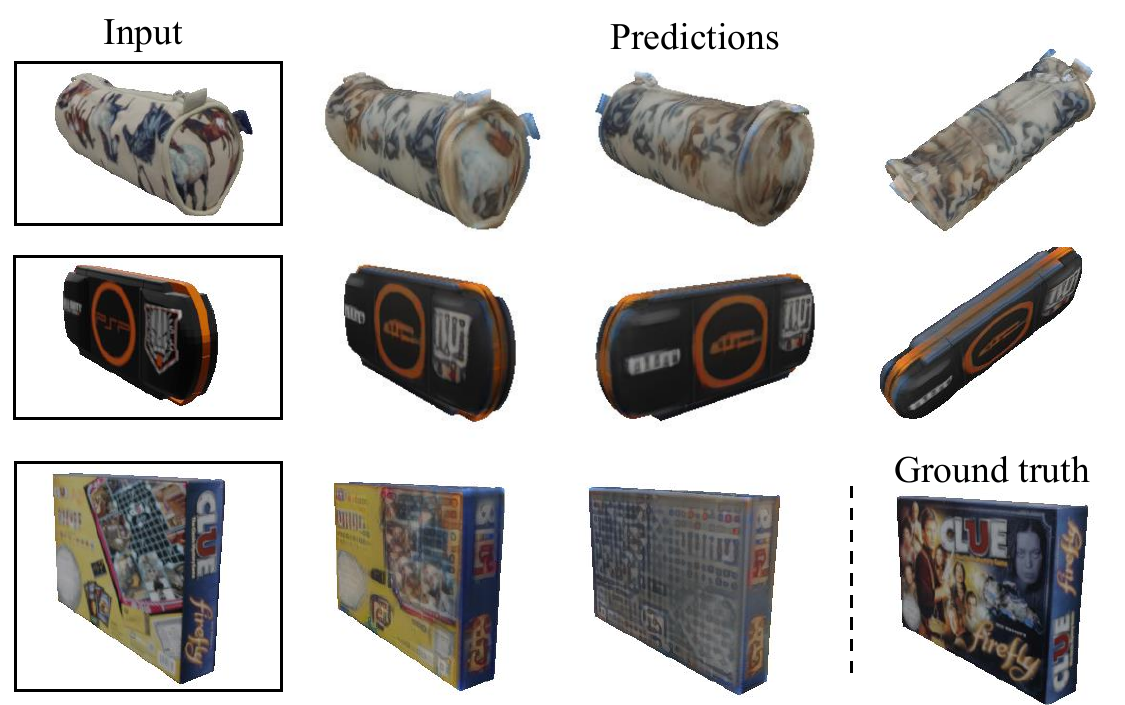}
    \caption{\textbf{Failure cases} While our method produces coherent textures in most cases, it struggles with high-frequency details. Typical failure modes include handling complex patterns (top row), reconstructing legible text (middle row), and recovering unseen regions such as the back of objects (bottom row).\vspace{-10pt}}
    \label{fig:failure_cases}
\end{figure}

While our method achieves strong results, it also has limitations. 
First, the output texture resolution is currently limited by the capacity of the model. 
As a result, our method can struggle to reproduce very fine details, such as text or high-frequency patterns (see Fig.~\ref{fig:failure_cases}). 

%% file: Sections/5_conclusion.tex


\section{Future work} 
Since our approach produces textures that are globally consistent and faithful to the input image, a promising direction is to incorporate a lightweight refinement stage that enhances high-frequency details. 
Another avenue is to integrate our feed-forward framework with generative pipelines (e.g., diffusion-based texturing), where our method could provide strong initialization and improve sample efficiency and fidelity. 
Finally, exploring generative extensions of our model would enable conditional sampling of multiple plausible texture fields for the same geometry, broadening its use in creative content generation.

\section{Conclusion}

We presented a transformer-based architecture for image-guided texture reconstruction that directly predicts continuous texture fields encoded using a triplane representation. 
Our method takes as input a single image and a mesh as input, does not rely on UV mapping or differentiable rendering, and generates high-quality textures in a single forward pass. 
Extensive experiments, ablations, and a user study demonstrate that our approach outperforms existing baselines in both fidelity and efficiency, making it a practical solution for large-scale 3D content creation.

\clearpage

%% file: main.bbl
\begin{thebibliography}{48}
\providecommand{\natexlab}[1]{#1}
\providecommand{\url}[1]{\texttt{#1}}
\expandafter\ifx\csname urlstyle\endcsname\relax
  \providecommand{\doi}[1]{doi: #1}\else
  \providecommand{\doi}{doi: \begingroup \urlstyle{rm}\Url}\fi

\bibitem[Bokhovkin et~al.(2023)Bokhovkin, Tulsiani, and Dai]{bokhovkin2023mesh2tex}
Alexey Bokhovkin, Shubham Tulsiani, and Angela Dai.
\newblock Mesh2tex: Generating mesh textures from image queries, 2023.

\bibitem[Cao et~al.(2023)Cao, Kreis, Fidler, Sharp, and Yin]{cao2023texfusion}
Tianshi Cao, Karsten Kreis, Sanja Fidler, Nicholas Sharp, and Kangxue Yin.
\newblock Texfusion: Synthesizing 3d textures with text-guided image diffusion models, 2023.

\bibitem[Chan et~al.(2021)Chan, Lin, Chan, Nagano, Pan, De~Mello, Gallo, Guibas, Tremblay, Khamis, et~al.]{chan2021efficient}
Eric~R Chan, Connor~Z Lin, Matthew~A Chan, Koki Nagano, Boxiao Pan, Shalini De~Mello, Orazio Gallo, LJ Guibas, J Tremblay, S Khamis, et~al.
\newblock Efficient geometry-aware 3d generative adversarial networks.” arxiv, 2021.

\bibitem[Chen et~al.(2023{\natexlab{a}})Chen, Siddiqui, Lee, Tulyakov, and Nießner]{chen2023text2tex}
Dave~Zhenyu Chen, Yawar Siddiqui, Hsin-Ying Lee, Sergey Tulyakov, and Matthias Nießner.
\newblock Text2tex: Text-driven texture synthesis via diffusion models, 2023{\natexlab{a}}.

\bibitem[Chen et~al.(2023{\natexlab{b}})Chen, Chen, Jiao, and Jia]{chen2023fantasia3d}
Rui Chen, Yongwei Chen, Ningxin Jiao, and Kui Jia.
\newblock Fantasia3d: Disentangling geometry and appearance for high-quality text-to-3d content creation, 2023{\natexlab{b}}.

\bibitem[Chen et~al.(2022)Chen, Chen, Lei, Zhang, and Jia]{chen2022tango}
Yongwei Chen, Rui Chen, Jiabao Lei, Yabin Zhang, and Kui Jia.
\newblock Tango: Text-driven photorealistic and robust 3d stylization via lighting decomposition, 2022.

\bibitem[Cheng et~al.(2023)Cheng, Li, Liu, and Wang]{cheng2023tuvf}
An-Chieh Cheng, Xueting Li, Sifei Liu, and Xiaolong Wang.
\newblock Tuvf: Learning generalizable texture uv radiance fields, 2023.

\bibitem[Deitke et~al.(2023)Deitke, Schwenk, Salvador, Weihs, Michel, VanderBilt, Schmidt, Ehsani, Kembhavi, and Farhadi]{deitke2023objaverse}
Matt Deitke, Dustin Schwenk, Jordi Salvador, Luca Weihs, Oscar Michel, Eli VanderBilt, Ludwig Schmidt, Kiana Ehsani, Aniruddha Kembhavi, and Ali Farhadi.
\newblock Objaverse: A universe of annotated 3d objects.
\newblock In \emph{Proceedings of the IEEE/CVF conference on computer vision and pattern recognition}, pages 13142--13153, 2023.

\bibitem[Dhariwal and Nichol(2021)]{DiffusionModels}
Prafulla Dhariwal and Alexander Nichol.
\newblock Diffusion models beat gans on image synthesis.
\newblock \emph{Advances in neural information processing systems}, 34:\penalty0 8780--8794, 2021.

\bibitem[Downs et~al.(2022)Downs, Francis, Koenig, Kinman, Hickman, Reymann, McHugh, and Vanhoucke]{downs2022google}
Laura Downs, Anthony Francis, Nate Koenig, Brandon Kinman, Ryan Hickman, Krista Reymann, Thomas~B McHugh, and Vincent Vanhoucke.
\newblock Google scanned objects: A high-quality dataset of 3d scanned household items.
\newblock In \emph{2022 International Conference on Robotics and Automation (ICRA)}, pages 2553--2560. IEEE, 2022.

\bibitem[Goodfellow et~al.(2014)Goodfellow, Pouget-Abadie, Mirza, Xu, Warde-Farley, Ozair, Courville, and Bengio]{goodfellow2014gans}
Ian~J. Goodfellow, Jean Pouget-Abadie, Mehdi Mirza, Bing Xu, David Warde-Farley, Sherjil Ozair, Aaron Courville, and Yoshua Bengio.
\newblock Generative adversarial networks, 2014.

\bibitem[Hong et~al.(2022)Hong, Zhang, Pan, Cai, Yang, and Liu]{hong2022avatarclip}
Fangzhou Hong, Mingyuan Zhang, Liang Pan, Zhongang Cai, Lei Yang, and Ziwei Liu.
\newblock Avatarclip: Zero-shot text-driven generation and animation of 3d avatars, 2022.

\bibitem[Hong et~al.(2024)Hong, Zhang, Gu, Bi, Zhou, Liu, Liu, Sunkavalli, Bui, and Tan]{hong2024lrm}
Yicong Hong, Kai Zhang, Jiuxiang Gu, Sai Bi, Yang Zhou, Difan Liu, Feng Liu, Kalyan Sunkavalli, Trung Bui, and Hao Tan.
\newblock Lrm: Large reconstruction model for single image to 3d, 2024.

\bibitem[Huo et~al.(2024)Huo, Guo, Zuo, Shi, Lu, Dai, Xu, Cheng, and Yang]{TexGen}
Dong Huo, Zixin Guo, Xinxin Zuo, Zhihao Shi, Juwei Lu, Peng Dai, Songcen Xu, Li Cheng, and Yee-Hong Yang.
\newblock Texgen: Text-guided 3d texture generation with multi-view sampling and resampling.
\newblock In \emph{European Conference on Computer Vision}, pages 352--368. Springer, 2024.

\bibitem[Karras et~al.(2019)Karras, Laine, and Aila]{karras2019stylegan}
Tero Karras, Samuli Laine, and Timo Aila.
\newblock A style-based generator architecture for generative adversarial networks, 2019.

\bibitem[Li et~al.(2023)Li, Tan, Zhang, Xu, Luan, Xu, Hong, Sunkavalli, Shakhnarovich, and Bi]{li2023instant3d}
Jiahao Li, Hao Tan, Kai Zhang, Zexiang Xu, Fujun Luan, Yinghao Xu, Yicong Hong, Kalyan Sunkavalli, Greg Shakhnarovich, and Sai Bi.
\newblock Instant3d: Fast text-to-3d with sparse-view generation and large reconstruction model, 2023.

\bibitem[Lin et~al.(2023)Lin, Gao, Tang, Takikawa, Zeng, Huang, Kreis, Fidler, Liu, and Lin]{lin2023magic3d}
Chen-Hsuan Lin, Jun Gao, Luming Tang, Towaki Takikawa, Xiaohui Zeng, Xun Huang, Karsten Kreis, Sanja Fidler, Ming-Yu Liu, and Tsung-Yi Lin.
\newblock Magic3d: High-resolution text-to-3d content creation, 2023.

\bibitem[Liu et~al.(2024)Liu, Wu, Liu, Liu, Wu, Peng, Zhao, Feng, Liu, and Ding]{liu2024texoct}
Jialun Liu, Chenming Wu, Xinqi Liu, Xing Liu, Jinbo Wu, Haotian Peng, Chen Zhao, Haocheng Feng, Jingtuo Liu, and Errui Ding.
\newblock Texoct: Generating textures of 3d models with octree-based diffusion.
\newblock In \emph{Proceedings of the IEEE/CVF Conference on Computer Vision and Pattern Recognition (CVPR)}, pages 4284--4293, 2024.

\bibitem[Ma et~al.(2023)Ma, Zhang, Sun, Ji, Wang, Jiang, Zhuang, and Ji]{ma2023xmesh}
Yiwei Ma, Xiaioqing Zhang, Xiaoshuai Sun, Jiayi Ji, Haowei Wang, Guannan Jiang, Weilin Zhuang, and Rongrong Ji.
\newblock X-mesh: Towards fast and accurate text-driven 3d stylization via dynamic textual guidance, 2023.

\bibitem[Michel et~al.(2021)Michel, Bar-On, Liu, Benaim, and Hanocka]{michel2021text2mesh}
Oscar Michel, Roi Bar-On, Richard Liu, Sagie Benaim, and Rana Hanocka.
\newblock Text2mesh: Text-driven neural stylization for meshes, 2021.

\bibitem[Mildenhall et~al.(2020)Mildenhall, Srinivasan, Tancik, Barron, Ramamoorthi, and Ng]{mildenhall2020nerf}
Ben Mildenhall, Pratul~P. Srinivasan, Matthew Tancik, Jonathan~T. Barron, Ravi Ramamoorthi, and Ren Ng.
\newblock Nerf: Representing scenes as neural radiance fields for view synthesis, 2020.

\bibitem[Mohammad~Khalid et~al.(2022)Mohammad~Khalid, Xie, Belilovsky, and Popa]{clipmesh}
Nasir Mohammad~Khalid, Tianhao Xie, Eugene Belilovsky, and Tiberiu Popa.
\newblock Clip-mesh: Generating textured meshes from text using pretrained image-text models.
\newblock In \emph{SIGGRAPH Asia 2022 Conference Papers}, page 1–8. ACM, 2022.

\bibitem[Müller et~al.(2022)Müller, Evans, Schied, and Keller]{instantngp}
Thomas Müller, Alex Evans, Christoph Schied, and Alexander Keller.
\newblock Instant neural graphics primitives with a multiresolution hash encoding.
\newblock \emph{ACM Transactions on Graphics}, 41\penalty0 (4):\penalty0 1–15, 2022.

\bibitem[Oechsle et~al.(2019)Oechsle, Mescheder, Niemeyer, Strauss, and Geiger]{oechsle2019texturefields}
Michael Oechsle, Lars Mescheder, Michael Niemeyer, Thilo Strauss, and Andreas Geiger.
\newblock Texture fields: Learning texture representations in function space, 2019.

\bibitem[Oquab et~al.(2024)Oquab, Darcet, Moutakanni, Vo, Szafraniec, Khalidov, Fernandez, Haziza, Massa, El-Nouby, Assran, Ballas, Galuba, Howes, Huang, Li, Misra, Rabbat, Sharma, Synnaeve, Xu, Jegou, Mairal, Labatut, Joulin, and Bojanowski]{dinov2}
Maxime Oquab, Timothée Darcet, Théo Moutakanni, Huy Vo, Marc Szafraniec, Vasil Khalidov, Pierre Fernandez, Daniel Haziza, Francisco Massa, Alaaeldin El-Nouby, Mahmoud Assran, Nicolas Ballas, Wojciech Galuba, Russell Howes, Po-Yao Huang, Shang-Wen Li, Ishan Misra, Michael Rabbat, Vasu Sharma, Gabriel Synnaeve, Hu Xu, Hervé Jegou, Julien Mairal, Patrick Labatut, Armand Joulin, and Piotr Bojanowski.
\newblock Dinov2: Learning robust visual features without supervision, 2024.

\bibitem[Perla et~al.(2024)Perla, Wang, Mahdavi-Amiri, and Zhang]{perla2024easi}
Sai Raj~Kishore Perla, Yizhi Wang, Ali Mahdavi-Amiri, and Hao Zhang.
\newblock Easi-tex: Edge-aware mesh texturing from single image.
\newblock \emph{ACM Transactions on Graphics (TOG)}, 43\penalty0 (4):\penalty0 1--11, 2024.

\bibitem[Poole et~al.(2022)Poole, Jain, Barron, and Mildenhall]{poole2022dreamfusion}
Ben Poole, Ajay Jain, Jonathan~T. Barron, and Ben Mildenhall.
\newblock Dreamfusion: Text-to-3d using 2d diffusion, 2022.

\bibitem[Qi et~al.(2017)Qi, Su, Mo, and Guibas]{qi2017pointnet}
Charles~R Qi, Hao Su, Kaichun Mo, and Leonidas~J Guibas.
\newblock Pointnet: Deep learning on point sets for 3d classification and segmentation.
\newblock In \emph{Proceedings of the IEEE conference on computer vision and pattern recognition}, pages 652--660, 2017.

\bibitem[Radford et~al.(2021{\natexlab{a}})Radford, Kim, Hallacy, Ramesh, Goh, Agarwal, Sastry, Askell, Mishkin, Clark, Krueger, and Sutskever]{radford2021clip}
Alec Radford, Jong~Wook Kim, Chris Hallacy, Aditya Ramesh, Gabriel Goh, Sandhini Agarwal, Girish Sastry, Amanda Askell, Pamela Mishkin, Jack Clark, Gretchen Krueger, and Ilya Sutskever.
\newblock Learning transferable visual models from natural language supervision, 2021{\natexlab{a}}.

\bibitem[Radford et~al.(2021{\natexlab{b}})Radford, Kim, Hallacy, Ramesh, Goh, Agarwal, Sastry, Askell, Mishkin, Clark, et~al.]{radford2021learning}
Alec Radford, Jong~Wook Kim, Chris Hallacy, Aditya Ramesh, Gabriel Goh, Sandhini Agarwal, Girish Sastry, Amanda Askell, Pamela Mishkin, Jack Clark, et~al.
\newblock Learning transferable visual models from natural language supervision.
\newblock In \emph{International conference on machine learning}, pages 8748--8763. PmLR, 2021{\natexlab{b}}.

\bibitem[Richardson et~al.(2023)Richardson, Metzer, Alaluf, Giryes, and Cohen-Or]{richardson2023texture}
Elad Richardson, Gal Metzer, Yuval Alaluf, Raja Giryes, and Daniel Cohen-Or.
\newblock Texture: Text-guided texturing of 3d shapes, 2023.

\bibitem[Rombach et~al.(2022)Rombach, Blattmann, Lorenz, Esser, and Ommer]{StableDiffusionModel}
Robin Rombach, Andreas Blattmann, Dominik Lorenz, Patrick Esser, and Bj{\"o}rn Ommer.
\newblock High-resolution image synthesis with latent diffusion models.
\newblock In \emph{Proceedings of the IEEE/CVF conference on computer vision and pattern recognition}, pages 10684--10695, 2022.

\bibitem[Saharia et~al.(2022)Saharia, Chan, Saxena, Li, Whang, Denton, Ghasemipour, Gontijo~Lopes, Karagol~Ayan, Salimans, et~al.]{Imagen}
Chitwan Saharia, William Chan, Saurabh Saxena, Lala Li, Jay Whang, Emily~L Denton, Kamyar Ghasemipour, Raphael Gontijo~Lopes, Burcu Karagol~Ayan, Tim Salimans, et~al.
\newblock Photorealistic text-to-image diffusion models with deep language understanding.
\newblock \emph{Advances in neural information processing systems}, 35:\penalty0 36479--36494, 2022.

\bibitem[Sanghi et~al.(2024)Sanghi, Khani, Reddy, Rampini, Cheung, Malekshan, Madan, and Shayani]{sanghi2024wavelet}
Aditya Sanghi, Aliasghar Khani, Pradyumna Reddy, Arianna Rampini, Derek Cheung, Kamal~Rahimi Malekshan, Kanika Madan, and Hooman Shayani.
\newblock Wavelet latent diffusion (wala): Billion-parameter 3d generative model with compact wavelet encodings.
\newblock \emph{arXiv preprint arXiv:2411.08017}, 2024.

\bibitem[Siddiqui et~al.(2022)Siddiqui, Thies, Ma, Shan, Nießner, and Dai]{siddiqui2022texturify}
Yawar Siddiqui, Justus Thies, Fangchang Ma, Qi Shan, Matthias Nießner, and Angela Dai.
\newblock Texturify: Generating textures on 3d shape surfaces, 2022.

\bibitem[Simonyan and Zisserman(2014)]{simonyan2014very}
Karen Simonyan and Andrew Zisserman.
\newblock Very deep convolutional networks for large-scale image recognition.
\newblock \emph{arXiv preprint arXiv:1409.1556}, 2014.

\bibitem[Tang et~al.(2024)Tang, Lu, Chen, Wen, Zeng, and Liu]{InTeX}
Jiaxiang Tang, Ruijie Lu, Xiaokang Chen, Xiang Wen, Gang Zeng, and Ziwei Liu.
\newblock Intex: Interactive text-to-texture synthesis via unified depth-aware inpainting.
\newblock \emph{arXiv preprint arXiv:2403.11878}, 2024.

\bibitem[Wang et~al.(2023)Wang, Lu, Wang, Bao, Li, Su, and Zhu]{wang2023prolificdreamer}
Zhengyi Wang, Cheng Lu, Yikai Wang, Fan Bao, Chongxuan Li, Hang Su, and Jun Zhu.
\newblock Prolificdreamer: High-fidelity and diverse text-to-3d generation with variational score distillation, 2023.

\bibitem[Xu et~al.(2024)Xu, Shi, Yifan, Chen, Yang, Peng, Shen, and Wetzstein]{xu2024grm}
Yinghao Xu, Zifan Shi, Wang Yifan, Hansheng Chen, Ceyuan Yang, Sida Peng, Yujun Shen, and Gordon Wetzstein.
\newblock Grm: Large gaussian reconstruction model for efficient 3d reconstruction and generation, 2024.

\bibitem[Yeh et~al.(2024)Yeh, Huang, Kim, Xiao, Nguyen-Phuoc, Khan, Zhang, Chandraker, Marshall, Dong, and Li]{yeh2024texturedreamer}
Yu-Ying Yeh, Jia-Bin Huang, Changil Kim, Lei Xiao, Thu Nguyen-Phuoc, Numair Khan, Cheng Zhang, Manmohan Chandraker, Carl~S Marshall, Zhao Dong, and Zhengqin Li.
\newblock Texturedreamer: Image-guided texture synthesis through geometry-aware diffusion, 2024.

\bibitem[Yu et~al.(2023)Yu, Dai, Li, Ma, Liu, and Qi]{yu2023pointuv}
Xin Yu, Peng Dai, Wenbo Li, Lan Ma, Zhengzhe Liu, and Xiaojuan Qi.
\newblock Texture generation on 3d meshes with point-uv diffusion, 2023.

\bibitem[Zamani et~al.(2025)Zamani, Xie, Aghdam, Popa, and Belilovsky]{TextureRewardLearning}
AmirHossein Zamani, Tianhao Xie, Amir~G Aghdam, Tiberiu Popa, and Eugene Belilovsky.
\newblock Geometry-{A}ware {P}reference {L}earning for 3{D} {T}exture {G}eneration.
\newblock \emph{arXiv preprint arXiv:2506.18331}, 2025.

\bibitem[Zeng et~al.(2023)Zeng, Chen, Qi, Liu, Zhao, Wang, Fu, Liu, and Yu]{zeng2023paint3d}
Xianfang Zeng, Xin Chen, Zhongqi Qi, Wen Liu, Zibo Zhao, Zhibin Wang, Bin Fu, Yong Liu, and Gang Yu.
\newblock Paint3d: Paint anything 3d with lighting-less texture diffusion models, 2023.

\bibitem[Zhang et~al.(2024{\natexlab{a}})Zhang, Pan, Zhang, Zhu, and Gao]{zhang2024texpainter}
Hongkun Zhang, Zherong Pan, Congyi Zhang, Lifeng Zhu, and Xifeng Gao.
\newblock Texpainter: Generative mesh texturing with multi-view consistency, 2024{\natexlab{a}}.

\bibitem[Zhang et~al.(2018{\natexlab{a}})Zhang, Isola, Efros, Shechtman, and Wang]{lpips}
Richard Zhang, Phillip Isola, Alexei~A. Efros, Eli Shechtman, and Oliver Wang.
\newblock The unreasonable effectiveness of deep features as a perceptual metric, 2018{\natexlab{a}}.

\bibitem[Zhang et~al.(2018{\natexlab{b}})Zhang, Isola, Efros, Shechtman, and Wang]{zhang2018unreasonable}
Richard Zhang, Phillip Isola, Alexei~A Efros, Eli Shechtman, and Oliver Wang.
\newblock The unreasonable effectiveness of deep features as a perceptual metric.
\newblock In \emph{Proceedings of the IEEE conference on computer vision and pattern recognition}, pages 586--595, 2018{\natexlab{b}}.

\bibitem[Zhang et~al.(2024{\natexlab{b}})Zhang, Liu, Xie, Yang, Liu, Yang, Zhang, Kou, Lin, Wang, and Jin]{zhang2024dreammat}
Yuqing Zhang, Yuan Liu, Zhiyu Xie, Lei Yang, Zhongyuan Liu, Mengzhou Yang, Runze Zhang, Qilong Kou, Cheng Lin, Wenping Wang, and Xiaogang Jin.
\newblock Dreammat: High-quality pbr material generation with geometry- and light-aware diffusion models, 2024{\natexlab{b}}.

\bibitem[Zou et~al.(2023)Zou, Yu, Guo, Li, Liang, Cao, and Zhang]{zou2023triplanemeetsgaussiansplatting}
Zi-Xin Zou, Zhipeng Yu, Yuan-Chen Guo, Yangguang Li, Ding Liang, Yan-Pei Cao, and Song-Hai Zhang.
\newblock Triplane meets gaussian splatting: Fast and generalizable single-view 3d reconstruction with transformers, 2023.

\end{thebibliography}
